# Descriptive evaluation of students using fuzzy approximate reasoning

Mohsen Annabestani[1], Alireza Rowhanimanesh[2], Aylar Mizani[3], Akram Rezaei[4]

1. Department of Electrical Engineering, Sharif University of Technology, Tehran, Iran.
2. Department of Electrical Engineering, University of Neyshabur, Neyshabur, Iran
3. Department of Curriculum Planning, Kharazmi University, Tehran, Iran.
4. Department of Computer engineering, Azad University of Mashhad, Mashhad, Iran.

**Abstract**

In recent years, descriptive evaluation has been introduced as a new model for educational evaluation of Iranian students. The current descriptive evaluation method is based on four-valued logic. Assessing all students with only four values is led to a lack of relative justice and creation of unrealistic equality. Also, the complexity of the evaluation process in the current method increases teacher error's likelihood. As a suitable solution, in this paper, a fuzzy descriptive evaluation system has been proposed. The proposed method is based on fuzzy logic, which is an infinite-valued logic and it can perform approximate reasoning on natural language propositions. By the proposed fuzzy system, student assessment is performed over the school year with infinite values instead of four values. But to eliminate the diversity of assigned values to students, at the end of the school year, the calculated values for each student will be rounded to the nearest value of the four standard values of the current descriptive evaluation system. It can be implemented easily in an appropriate smartphone app, which makes it much easier for teachers to assess the educational process of students. In this paper, the evaluation process of the elementary third-grade mathematics course in Iran during the period from the beginning of the MEHR (The Seventh month of Iran) to the end of BAHMAN (The Eleventh Month of Iran) is examined by the proposed system. To evaluate the validity of this system, the proposed method has been simulated in MATLAB software.

## I- INTRODUCTION

Faculty performance evaluation is an all the time challenging issue [1]. Especially in the developing nations, where instead of any scientific and standardized assessment methods the evaluation mostly depends on the qualitative judgment of the academic administrators[1]. Human as a system is very complex and descriptive assessment of students is a nonlinear, complex and uncertain process which depends on several parameters. To control this human-based system, modeling of human intelligence can be a proper candidate. In the student descriptive evaluation system of Iran as a case study, teachers can use all of their impressions and self-results about students in the process of education into a form of qualitative adjectives and phrases. In the current system of descriptive evaluation in Iran at the end of each semester and also at the end of school year, teachers describe the educational situation of students by some linguistic terms using the average of their recorded qualitative history through the semester or school year. It is obvious that using linguistic variables are more useful than numeric feedbacks like Grade Point Average (GPA) and this method makes students more relax and stressless during education. Hence they can more focus on learning instead of competition to catch better scores. Therefore for a true, real, complete and fair evaluation of students, we can't avoid the descriptive and qualitative feedbacks, especially in the elementary schools. When we work with human approaches and linguistic variables, using fuzzy theory would be one of the best methods. Fuzzy logic success is for its capability in describing system behaviors by simple conditional IF-THEN rules. In the most applications, this capability provides a simple solution which needs to spend a short time to design the system. In addition, all information and knowledge relating to the desired system for optimization and having better performance will be used directly[2]. We can find a variety of applications for fuzzy theory, for example in [3-5] using Adaptive Neuro-Fuzzy Inference System (ANFIS) a nonlinear identification method has been proposed, or in [6-8] they have designed a fuzzy criterion in order to measure the quality of step response. Or we can find other fuzzy applications in the fields of Image processing[9, 10], wind turbine [11], stock exchange[12], control[13], etc.

Traditional grading methods are significantly based on human judgments, which tend to be subjective [14]. In addition, they are based on crisp criteria which make a proper condition for assigning erroneous scores. Hence it can be a potential source of uncertainties which might impair the credibility of the evaluation system. In order to reduce the uncertainties and providing more objective, reliable, fair and precise grading; several fuzzy based methods have been developed. For example, Echauz and Vachtsevanos presented a fuzzy-based grading system that utilizes student's and instructor's performance measures in order to modify a set of collectively approved, a priori fuzzy grades, so as to produce a "fair" mark distribution [15]. In another work, a fuzzy set approach was proposed in order to the assessment of student-centered learning. The proposed fuzzy set approach incorporates student's opinions into assessment and allows them to have a better understanding of the assessment criteria [16]. Toward to cover more uncertainties, the type-2 (T2) fuzzy set was used in some approaches. For example in [17], an interval type-2 (IT2) fuzzy sets, which are a special case of the general T2 fuzzy sets, was used. The transparency and capabilities of type-2 fuzzy sets in handling uncertainties are expected to provide an evaluation system able to justify and raise the quality and consistency of assessment judgments[17]. Jamsandekar and Mudholkar performed the performance evaluation of students based on fuzzy inference technique. They proposed an approach which was a combination of two membership functions. The fuzzy approach was further compared with traditional methods for evaluating the variance [18]. Yadav and Singh proposed a fuzzy expert system for evaluation of student academic performance. They also proposed several approaches using fuzzy logic techniques to provide a practical method for evaluating student's performance and comparison with existing statistical methods [19]. Jyothi et al. proposed a fuzzy expert system for evaluating teachers overall performance based on fuzzy logic techniques[20]. Nunes and Neill described an experiment where team performance was evaluated using fuzzy logic reasoning approach. The results showed that intelligent fuzzy controllers were able to perceive and evaluate the Team's performance [21]. Ingoley and Bakal proposed a fuzzy based system which considers vagueness of question paper beside accuracy rate, complexity, and importance. The proposed technique provides more transparent and fairer results to all students [22]. Yadav et al. presented a new fuzzy expert system (NFES) for performance evaluation of students[23]. Yildiz and Baba proposed a new approach based on fuzzy decision support systems for the evaluation of student's performance. Their model was based on a fuzzy multi-criteria method for evaluating student's performance in laboratory activities. The results showed better performance of fuzzy systems over classical systems [24]. Kharola and Kunwar proposed a new methodological approach using fuzzy logic reasoning for performance evaluation of students [25]. All above-mentioned papers are related to the use of fuzzy logic in the academic evaluation, but not specific research can be found specifically for its application in the descriptive evaluation.

The current version of descriptive evaluation in Iran is based on the teacher's perception from student's activity means that the teacher expresses the status of students using qualitative phrases [26]. Currently, in Iran these qualitative phrases are four phrases: "Need More Effort (**NME**)," "As Expected (**AE**)," "Good (**G**)" and "Very Good (**VG**)." Now there are some questions here, is it possible to describe this complex process only by four qualitative phrases (four-valued logic)? It means that, are the students just in the state of need to have more effort (NME), or as expected (AE), good (G) or very good (VG)? Do the students may not be situated between a case of good (G) and very good (VG) or between good (G) and as expected (AE)? The answer to these questions is clear. Never can fully and fairly describe the students using these four qualitative phrases. For example if the numerical equivalent of Very Good (VG) is considered around 18 to 20 , definitely the student who have gotten score 20 are belong more to the set of Very Good students (VG) in comparison to students who have reached 18, or on the other hands both of them are very good but 20 is better than 18, or even a student who is gotten score18, but his educational performances during the school year change him to or even better than the student who has reached 20 and this is very good. Unfortunately, the current methods do not distinguish between these unequal cases, and all of them are shown at the same level "Very Good." The teacher knows these differences, but because of the weaknesses in tools and methods of description and the complexity of the descriptive evaluation process, the teacher has to use this method. The rest of the paper has been categorized into three parts, in part II we will mathematically justify the problem. Our main achievement in this paper, i.e., Fuzzy Descriptive Evaluation System (FDES) will be described in part III, and finally, in part IV, we will discuss on the results.

## II- FUZZY LOGIC IN DESCRIPTIVE EVALUATION PROCESS

If we assume $U$ is the universe, a collection of objects denoted generically by $x$, a set of ordered pairs as (1) represents a fuzzy set like $A$ in the universe $U$ [27]:

$$A = \{(x, \mu_A(x)) / x \in U\} \qquad (1)$$

Where, $\mu_A(x) \in [0,1]$ is a fuzzy membership function which maps $U$ to a membership space. Therefore, a fuzzy set is the generalization of a classic set that allows the members of each set have any value in the interval of [0,1]. In other words, a classic set can only assign two values of 0 and 1 to its members, while a fuzzy set using a membership function can assign to its members any values of the range [0,1] [27, 28]. If we want to connect the fuzzy concept to descriptive evaluation, we should say that, in the current educational system of Iran, the numerical range between 10 to 20 are divided into mentioned four qualitative phrases, NME, AE, G, and VG. It means that for example 16, 16.3 and 17.2 all are good, but the reality is different, and 17.2 is better than the others. This flaw is due to a four-valued definition of qualitative phrases. We should define this set in a range (e.g., 10 to 20 or zero to 1) that there would be infinite valued instead of four-values. To achieve these changes we should use fuzzy sets like Fig.1 instead of the classic sets.

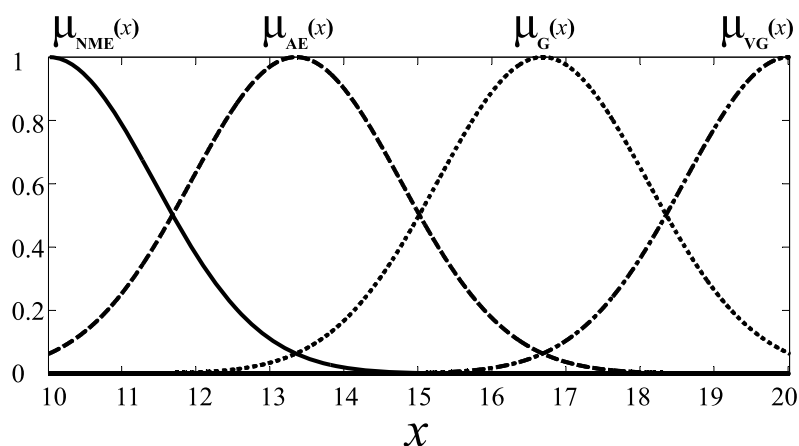

Fig. 1: Symmetrical Gaussian functions

Four values NME, AE, G, and VG, are turning points of all infinite values and they have continues overlap, for example, a good student with more efforts can reach to a very good student level and in this process experiences some cases between good and very good.

### III - FUZZY DESCRIPTIVE EVALUATION METHOD

In the current descriptive evaluation method of Iran, each course is divided into several indicators, and from the beginning of MEHR(The Seventh month of Iran), these indicators will be taught to students respectively. The assessment of previous indicators will be continued by adding each new branch, assessing each branch at any time should be considered the status of the student in previous assessment stage of the same indicator. This process for all of the indicators and even the result of compounding these indicators will be continued until the end of the school year. This complexity will make the assessment and evaluation hard for the teacher and then make the human error. In the following list, all of the evaluation indicators for the mathematics lesson in the elementary third grade during the period of MEHR to end of BAHMAN have been listed:

A. *Proper skill in reading, writing and comparing numbers.*
B. *Getting to know with measurement units and calculations.*
C. *Calculating and drawing geometric shapes.*
D. *Ability to perform calculations and four basic operations.*
E. *Ability to express the information and making the correct conclusions using the graph*

As mentioned earlier, the target of this paper is designing a fuzzy descriptive evaluation system (FDES) that as illustrated in Fig.2, parameters A, B, C, D and, E are inputs of this system. Each of these parameters has a qualitative value at each stage of the evaluation, which is a function of the previous evaluations.

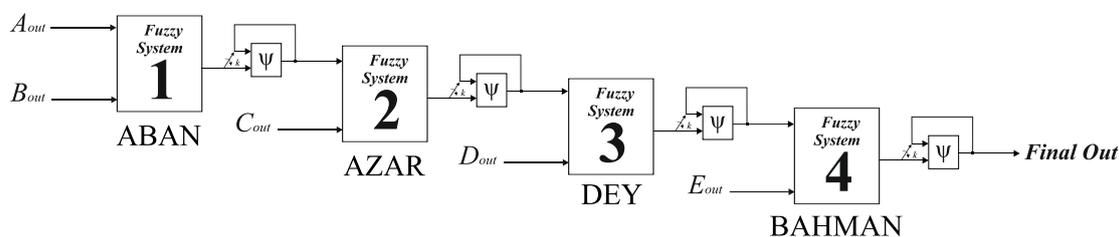

**Fig.2: Schematic of the fuzzy descriptive evaluation system (FDES).**

For example in parameter $A$, if at the last step of the evaluation we call it $A_{Out}$, and evaluate the related parameters of this parameter in the whole process, as $A_{m1}, A_{m2} ... A_{mn}$, we will have (2):

$$A_{Out} = f_A(A_{m1}, A_{m2}, A_{m3}, ..., A_{mn}) \quad (2)$$

Where $f_A$ describe the relationship between the parameters as a nonlinear and causal system using a hierarchical network of $n-1$ fuzzy systems as shown in Fig.3:

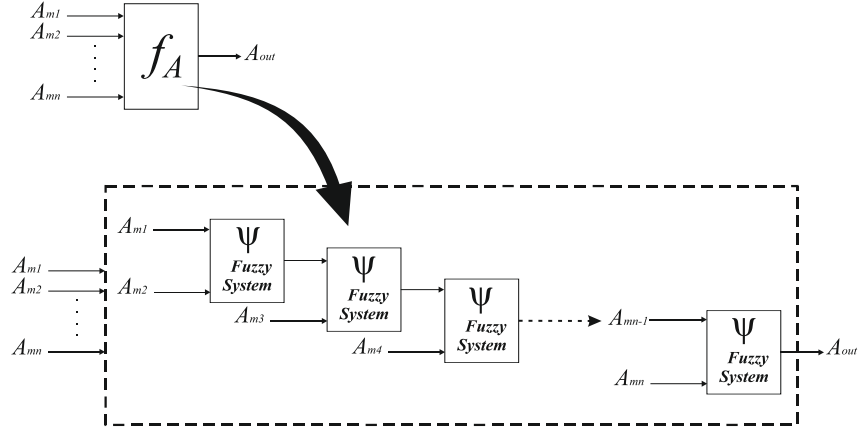

Fig. 3: A hierarchical network of $n-1$ the fuzzy system as $f_A$.

In the above system, $\psi$ is a two-input, and single-output fuzzy system, which its inputs are the derived statements from two consecutive evaluations and its output is the valued-mean of the inputs. The meaning of the valued-mean is that the value and effect of the input statements are also applied to the logical outcome of the output. A hierarchical network of $n$-$1$ $\psi$ fuzzy system for one of the evaluation indicators makes a system like $f_A$ for the realization of equation (2). Like $f_A$, we will have four other systems $f_B$, $f_C$, $f_D$ and $f_E$ that respectively calculate the indicators $B_{out}$, $C_{out}$, $D_{out,}$ and $E_{out}$.

## A. Using feedback instead of the hierarchical structure

As depicted in Fig.3, the system $f_A$ is created using $n$-$1$ fuzzy system ($n$ is the number of input propositions). But as an alternative for this system, we can incorporate proper feedback signals and propose a closed loop system like Fig.4.

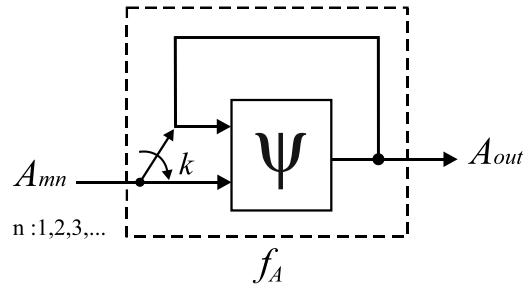

Fif.4: The feedbacked version of $f_A$ the system.

In this system, the key $k$ is closed only at $n = 1$ and works as follows:
First, at $n = 1$, the first proposition of evaluation ($A_{m1}$) is applied to both inputs of the system, which the result will be almost the same proposition ($\approx A_{m1}$). At $n = 2$, by applying the second evaluation proposition ($A_{m2}$) to the second input, the result of the previous one ($\approx A_{m1}$) is feedbacked to the first input. That is, using this feedback, the first proposition ($A_{m1}$) and the second one ($A_{m2}$) are combined. Like the mentioned procedure at $n = 1$, the result is feedbacked again to the first input and combined with the third evaluation proposition ($A_{m3}$). Similarly, this process will be proceeds to stage $n$. To prove the advantage of this feedbacked system instead of the open loop ones we should say that, assessment is a time-variant process and maybe a teacher want to have several evaluations (by quiz, oral question, homework, and any other feedbacks) about each of evaluation indicators (A, B, C, D and E) through the time that he or she is teaching. But the proposed FDES system get the valued-mean as input, and a hierarchy of the $\psi$ fuzzy systems make these valued-means ($A_{out}$ to $E_{out}$). The closed-loop system that we have presented here help the teacher to add his/her feedbacks about each of the evaluation indicators (*A, B, C, D,* and *E*) in any moment of school year and since we don't know how many times the teacher wants to assess the evaluation indicators, we cannot design an opened loop system like the ones that has been depicted in Fig.3.

### B.  $\psi$ Fuzzy system

As depicted in Fig.5, the $\psi$ fuzzy system has two inputs. In this system, the inputs are the statements derived from the evaluation of the stage *m* ($X_m$), and the stage *m + 1* ($X_{m+1}$) and its output ($X_{Out}$) is the valued-mean of the inputs. To design the membership functions of the $\psi$ fuzzy system we have used a set of Gaussian functions like Fig.1 for both inputs as well as output. Since the type and interval of definition of inputs and outputs are similar, it is meaningful enough to choose similar membership functions.

### C1. Inference of $\psi$ fuzzy system

In the inference step, 16 IF-THEN fuzzy rules are defined (Table.1). In these rules, the product T-norm operator is used. The reason for using T-norm is that the fuzzy conditional statements of the knowledge base are not separated from each other and all refer to a single truth. The reason for using the product T-norm is the effect of the entire range of membership functions as well as the continuity of the output. In the stage of fuzzy inference, we have used the Mamdani-product inference machine. That is, we first combine the rules, and then the input of the knowledge base applies to the result of this combination [29]. In Fig.5, the relationship between both inputs and the output of the $\psi$ fuzzy system has been shown as a three-dimensional surface. You will see that this surface is in accordance with the knowledge base of this system.

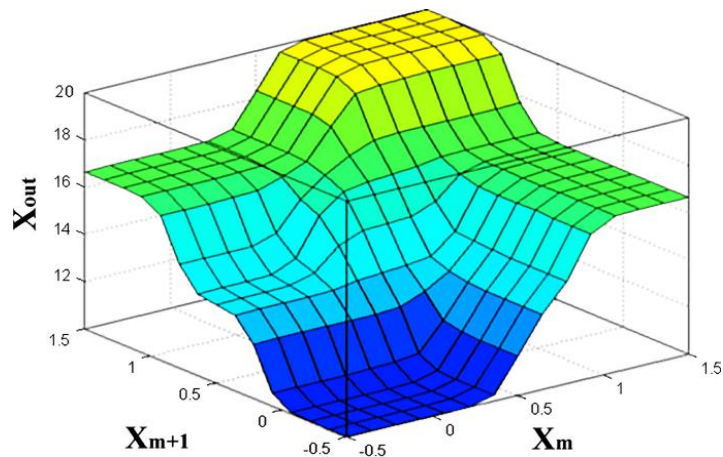

**Fig.5: the 3D surface of the $\psi$ fuzzy system**

**Table.1: IF-THEN rule table of $\psi$ fuzzy system**

|            | NME(m) | AE(m) | G(m) | VG(m) |
|------------|--------|-------|------|-------|
| **NME(m+1)** | NME    | NME   | AE   | G     |
| **AE(m+1)**  | AE     | AE    | AE   | G     |
| **G(m+1)**   | AE     | G     | G    | G     |
| **VG(m+1)**  | G      | G     | VG   | VG    |

## C2. Simulation of $\psi$ fuzzy system

In the $\psi$ system, in most of the cases, the importance of the second input ($X_{m+1}$) is more than the first one ($X_m$), and this difference is attributed to the fact that the student's progress and failures are more important in subsequent evaluations. Of course, this view is only discussed in the $\psi$ system, and in systems 1, 2, 3, and 4, the primary attention is paid to previous evaluations, which this view also points out to the fact that based on the continuity of mathematical course, it is expected that the students first learn the previous lessons well. To verify the system's performance, we have simulated the proposed method in the MATLAB R2010a software and, by creating the simulated data for 30 days, it is demonstrated that the system works well with the teacher's knowledge and vision.

As described in the description of the $\psi$ system, we expect the output to be almost equal to the inputs at the time of relative equilibrium of Inputs. By looking at the dashed line curves in the simulation figure (Fig. 6), you can see the system works well in this state and all curves ($X_m$, $X_{m+1}$ and $X_{Out}$) fluctuate around 14. Now, we want to show that the importance of the second input ($X_{m+1}$) is higher than the first input ($X_m$). Take a look at the dotted line samples of the simulation. You will find that the first input ($X_m$) is fluctuating around a very good term (VG), and the second input ($X_{m+1}$) is around the good statement (G).

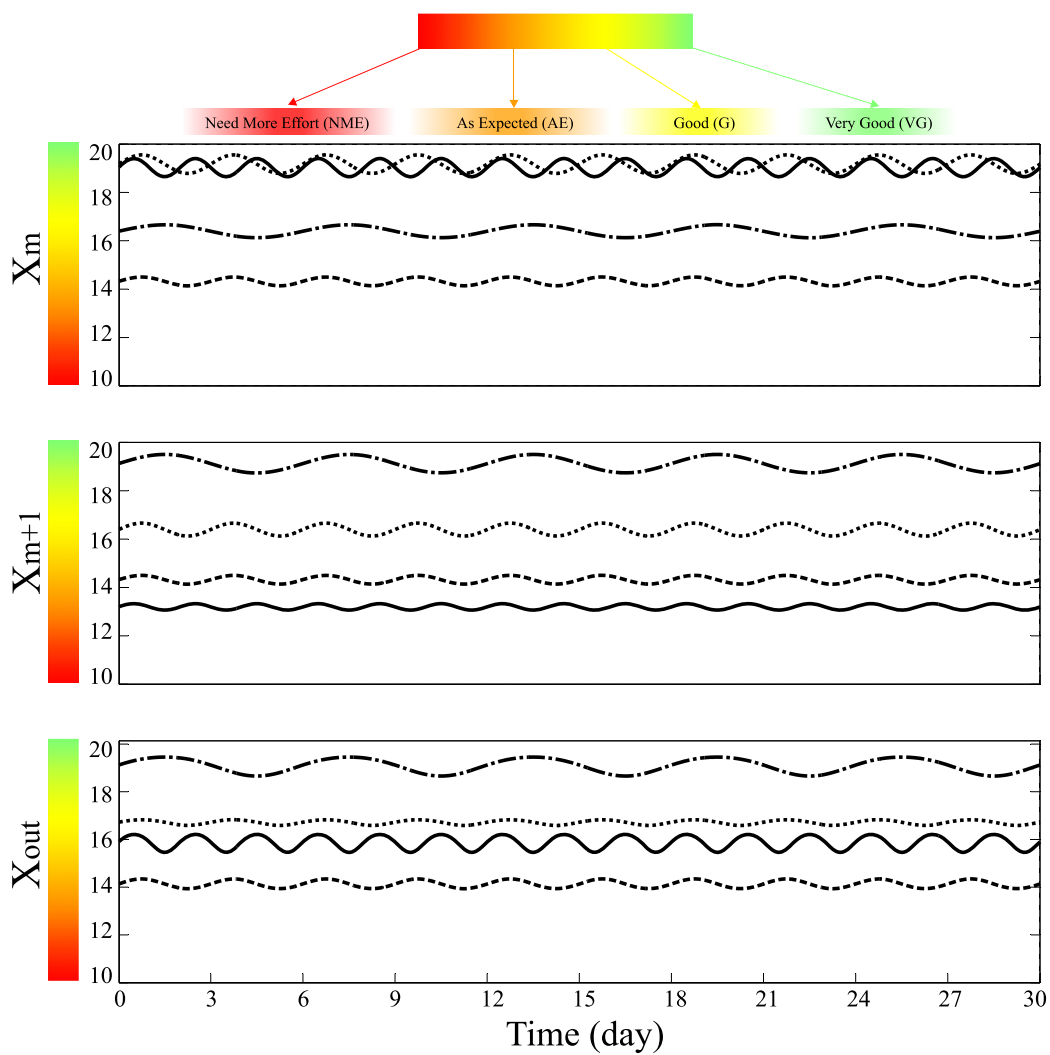

**Fig.6: The simulation results of $\psi$ fuzzy system**

Regarding the fuzzy rule base (Table.1), we expect the output ($X_{Out}$) to be around good too, but since the second input ($X_{m+1}$) is very good, its goodness is a bit better than the goodness of first input ($X_m$). In Fig.8 you can see that this expectation is well-fulfilled. Let's look at the dash-dotted line samples of the simulation image. The inputs of this sample are the contrariwise of the inputs of the dotted sample, that is, the first input ($X_m$) fluctuates around the good (G), and the second input ($X_{m+1}$) is located around the very good statement (VG), according to the fuzzy rules base of $\psi$ system (Table.1), we expect that the output ($X_{Out}$) should be around very good (VG), but because of the goodness of the first input ($X_m$) the power of this very good result is a bit lower than the power of very good result of the second input ($X_{m+1}$). You can see that this expectation is also well-fulfilled. Finally as shown by solid line samples, the first input ($X_m$) is fluctuating around a very good term (VG) and the second input statement ($X_{m+1}$) is near the As Expected range (AE).

Regarding the fuzzy rule base, we expect the output ($X_{Out}$) fluctuates around the good statement (G), but a good statement that its goodness is lower than the goodness of dotted line output, since in dotted one the good statement has been deducted from the combination of good (G) and very good (VG) but here in the solid line one the output resulted from good (G) and As Expected (AE). Hence we can find this is a natural reduction in the power of good statement. You can find all of these facts in the results of Fig.6. Finally, by observing all of the examples, we will always see that the second input ($X_{m+1}$) is dominant, which also confirms the correctness of the system's performance.

## C. FDES system

As depicted in Fig.7 the Fuzzy Descriptive Evaluation System (FDES) consists of four main sub-systems which are called 1, 2, 3 and 4 that their outputs through $\psi$ fuzzy systems are compared with previous samples in a comparable process, and in every second a signal is produced as the first input of the next main system.

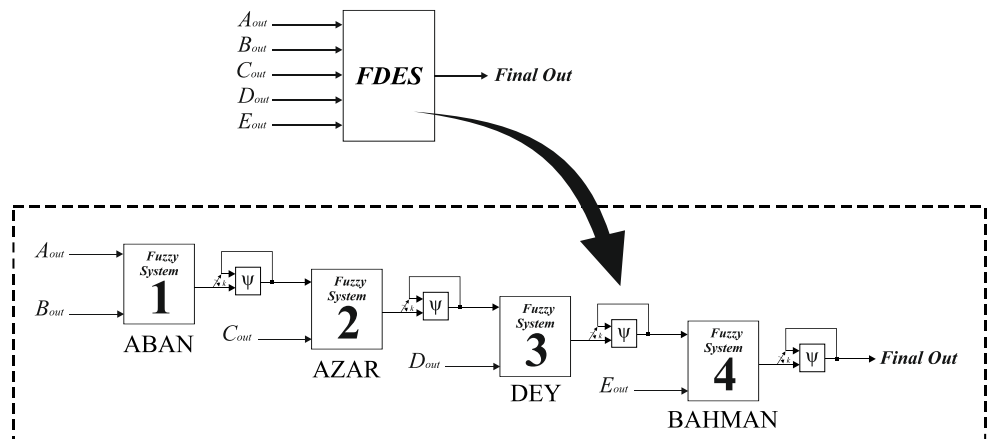

**Fig.7: FDES Structure.**

Here we have four outputs that each of them is a function of previous outputs and inputs and *"Final Out"* is a function of all previous outputs and evaluation indicators. At the end of the school year *"Final Out"* will be reported as the evaluation result of students. *"Final Out"* is a general descriptive phrase, but to reduce the diversity of assigned qualitative phrases to students, it will be rounded to one of four standard qualitative phrases, i.e., NME, AE, G, and VG.

## C1. Hierarchical structure of FDES

Hierarchical structures are well-known structure to solve verity of problems. They are various and we can find them in different topics such as estimation of network's missing links [30], organization of genetic in cortical surface area [31], primate cerebral cortex processing [32], analysis of group dynamics in pigeon flocks [33],designing algorithms for force calculation [34], nonlinear and non-autoregressive identification of IPMC soft actuators [4] etc. Before explaining the FDES system, we want to explain the reason behind using hierarchical structure and its division to four subsystems. The reason for dividing the FDES system into four subsystems is the increasing of system dimension and therefore increasing the number of its fuzzy rules. If we assume there are *m* input variables and *n* fuzzy sets are defined for each input variable, then the number of fuzzy rules will be $n^m$ that for large *m*, $n^m$ would be a very large number. The number of rules in the FDES system becomes $n^m = 4^5 = 1024$ that this high number of rules is a major problem. To resolve this problem, we have used the hierarchical structure to connect four fuzzy systems. Each fuzzy system with a low dimension (low rules) is a level established in a hierarchical fuzzy system. If *m* input variable exists ( $x_1, x_2,....x_m$ ) then the levels of hierarchical systems are composed as the following [28]:

**The first level:** a fuzzy system would be with $m_1$ input variables ( $x_1, x_2,....x_{m_1}$ ) that will be made by the following rules:

$$if \ x_1, A_1^w \ And \ ...... \ And \ x_{m_1}, A_{m_1}^w \ Then \ B_1^w \qquad (3)$$

Where

$$w = 1, 2,....., M_1 \qquad 2 \le m_1 < m$$

**The k level (k>1):** a fuzzy system with $m_k + 1$ input variables ( $m_k \ge 1$ ) will be made by the following rules:

$$if \ x_{N_k+1}, A_{N_k+1}^w \ And \ ...... \ And \ x_{N_k+m}, A_{N_k+m}^w \ And \ y_{k-1}, C_{k-1}^w \ Then \ y_k, B_k^w \qquad (4)$$

$$w = 1, 2,....., M_k \qquad N_k = \sum_{j=1}^{k-1} m_j \qquad (5)$$

It will be continued to $k = w$ and then $\sum_{j=1}^{k-1} m_j = m$ , means all the input variables used in one of the levels. In the *FDES* system ( $m_1 = 2$ and $m_k = 1$ ), all fuzzy systems of the hierarchical system have two inputs, and the number of levels will be *w = m-1*. It is proved that the number of rules in a hierarchical fuzzy system is a linear function of the number of input variables and the number of rules in case of two-inputs (Fig.8) will be reached to the minimum numbers of rules [28].

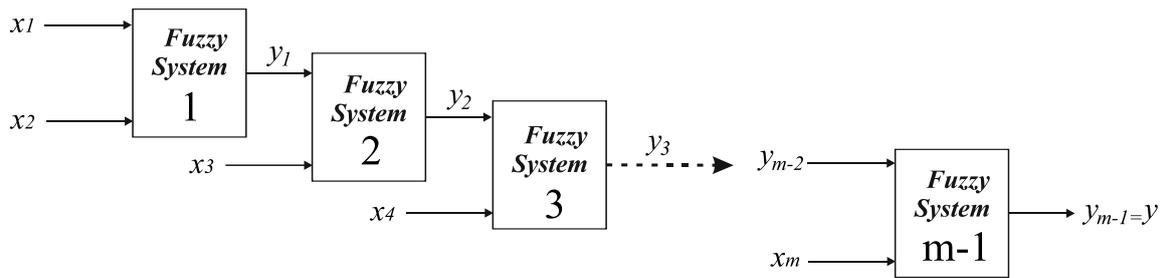

**Fig.8: The hierarchical structure in case of two-input levels.**

## C2. Simulation of FDES

By designing a simple smartphone app, the teacher at any time of the school year can have a logical inference from the student's activities, whereby the system provides to have a complete report of student learning activities. The teacher has complete knowledge of each of the relevant educational indicators for each student. Understand their progress, weakness, strength, and abilities at every time about any of the indicators and even a combination of indicators, and can make the most important decisions about student's education using this comprehensive assessment. At the end of the school year, that here it has been assumed the end of BAHMAN, the system presents an educational record for each lesson, in which the student's overall and final status along with the status of each of the indicators, is provided analytically. In addition, the report provides a qualitative statement as the student's grade in each course. Here to evaluate the performance of FDES approach the evaluation process of the elementary third-grade mathematics course in Iran during the period from the beginning of the MEHR to the end of BAHMAN has been examined for a typical student. As mentioned before there are five evaluation indicators (A, B, C, D, and E) for the mathematics course in the elementary third grade in Iran that we've defined them here as five continues time variant functions (Figs 9-13). By applying these functions to the FDES system, the assessment result of the mentioned typical student will be obtained as Fig.14. In this part, we want to describe all of these assumed indicators and then talk about the result and the performance of the system.

*Indicator A:* *Proper skill in reading, writing and comparing numbers.*

Reading and writing skills and comparing of numbers are good at the beginning of the month and start with the value between "**good**" and "**very good**", and progressively progressing, so that in the period between AZAR(The ninth month of Iran) and DEY(The tenth month of Iran), this indicator is in its highest level, and after that it returns and falls gradually to a value between "**good**" and "**very good**" at the end of the BAHMAN.

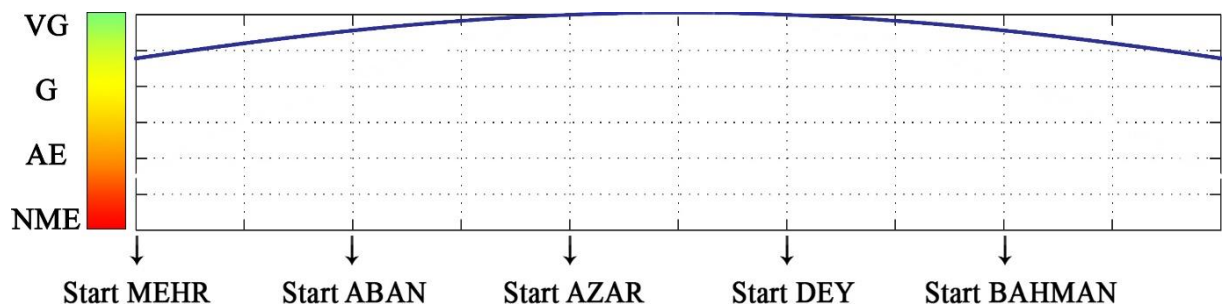

Fig. 9: Educational process of indicator *A* for a typical student.

*Indicator B:* *Getting to know with measurement units and calculations.*

From the beginning of ABAN (The seventh month of Iran), the student begins to work in an "**almost good**" position and continues to grow increasingly, reaching the highest level of "**very good**" at the end of the BAHMAN.

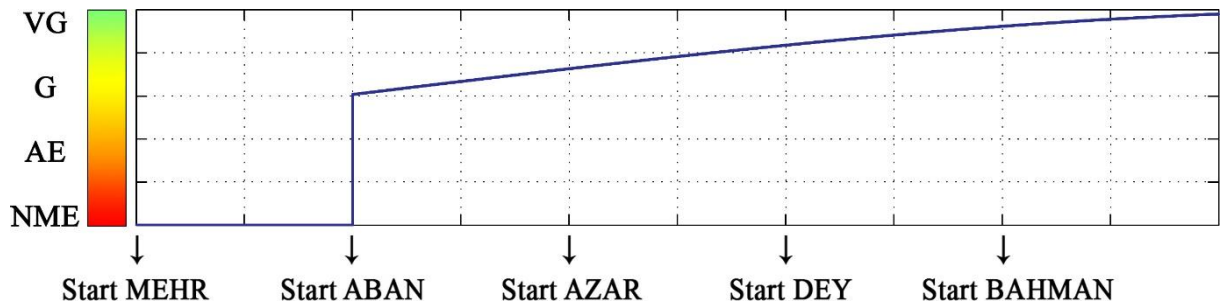

**Fig. 10: Educational process of indicator B for a typical student.**

*Indicator C*: *Computing and drawing geometric shapes.*

This indicator, which is measured from the start of AZAR is "**almost very good**" till the beginning of the DEY, but after that, it begins to fall and, finally, at the end of the BAHMAN it becomes to an **"almost weak but tolerable"** student.

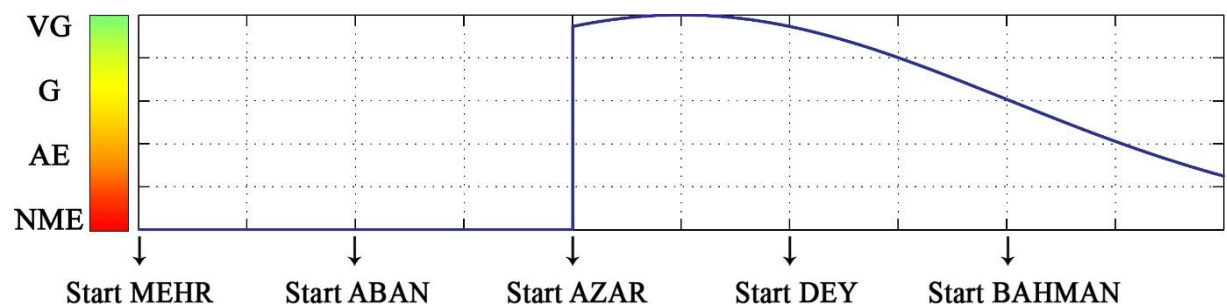

**Fig. 11: Educational process of indicator C for a typical student.**

*Indicator D*: *Ability to perform calculations and four basic operations.*

This indicator is evaluated from the beginning of the DEY. So it is initially in a "**good**" state, and over the time, with limited fluctuations, it progresses ascending, and at the end of BAHMAN, it is reached to the best possible level of "**very good**."

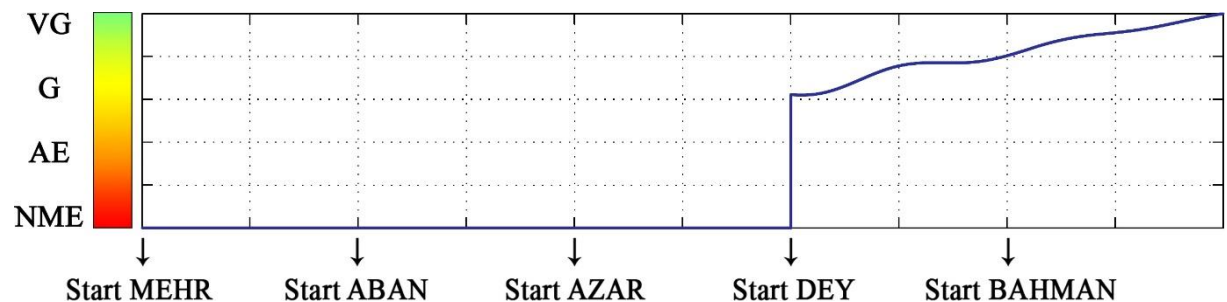

**Fig. 12: Educational process of indicator *D* for a typical student.**

*Indicator E*: *Ability to express the information and making the correct conclusions using the graph*

This indicator is oscillating. At the beginning, it is in **"a range between good and very good,"** which is more close to good. Then it slows down a bit in a short time but starts to progress again and in the middle of the BAHMAN reached the best possible level (**very good**), but it begins a slowly decreasing and symmetrically continues with the first half of the period till the end of BAHMAN.

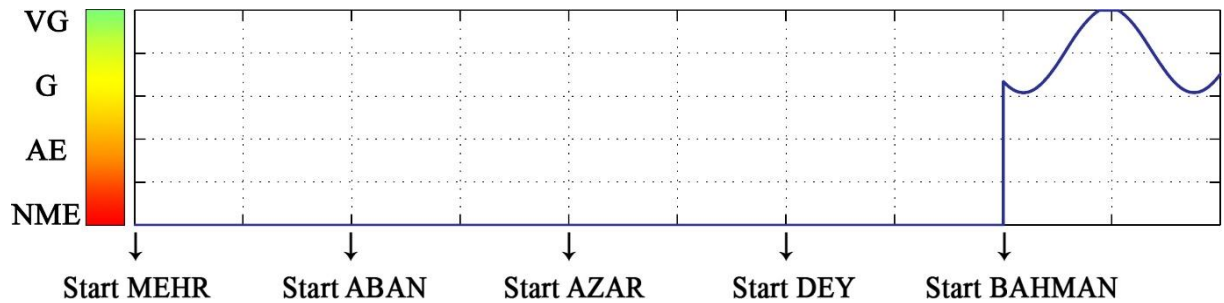

**Fig. 13: Educational process of indicator *E* for a typical student.**

Now we want to have qualitative feedback from the student's activity at each moment in the school year due to all of the indicators (A, B, C, D, and E). The graph in Fig.16 shows this qualitative feedback as a continuous function throughout the beginning of the MEHR to the end of BAHMAN. For example, at the point $\alpha_1$, which is near the end of ABAN, the assumed student is in **"good"** condition. At the end of AZAR ($\alpha_2$) is in a **"range between good and very good"** and at the middle of BAHMAN ($\alpha_3$), it is very close to good. Using this system in each moment of a year, the teacher can have a logical assessment influenced by all the indicators affecting the evaluation process.

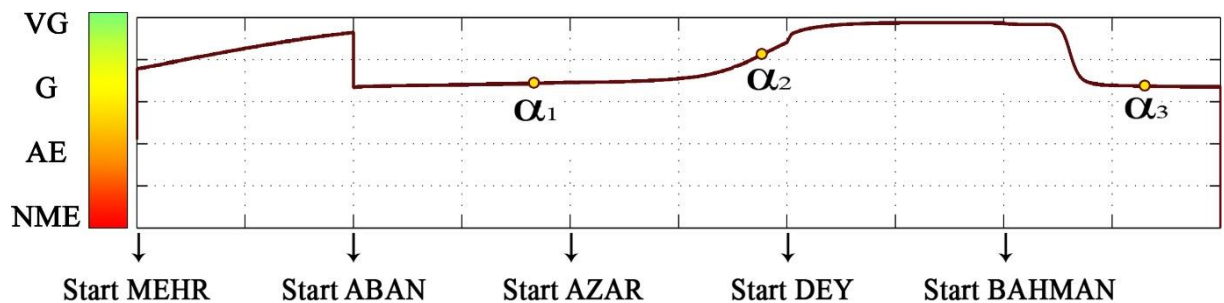

**Fig.14: Educational process of a typical student from the beginning of MEHR until the end of BAHMAN which resulted from all evaluation indicators (A, B, C, D, and E).**

## IV - CONCLUSION

In this paper, we modeled the teacher's knowledge to eliminate drawbacks in the current descriptive evaluation system of Iran (four-valued logic) in a logical and mathematical mechanism with the help of fuzzy logic (infinite-valued logic). We can also use this human model in analyzing and evaluating the educational process of students electronically using a simple smartphone app. By observing the simulation results in FDES, we found that in every time of the school year, we can have a logical inference of the student's performances in each evaluation indicators and the result of combining these indicators. Each evaluation by considering effects of previous evaluations has its effect in the result of the evaluation propositions and creates a real and fair assessment, eliminates human error, and looks wisely to student's changes, and offers comprehensive feedback of students in natural human language. By simulating this system in the MATLAB R2010a environment, its performance was validated, and the system's compliance with the teacher's perspective was confirmed.

Finally, it is mentioned that the using of fuzzy logic in the evaluation process will not make the evaluation more difficult and more complex. This method by reducing the human error and increasing the accuracy in measuring the specified parameters in the evaluation process of student makes a more realistic and more reliable result because the students are evaluated over the school year with infinite values (Realistic) instead of four values (Unrealistic). Also, this system is implementable as a simple smartphone app; for this reason, the evaluation process will become simpler for teachers. It is very important to say that the main reason to design descriptive evaluation as an alternative for the score-based method is avoiding to categorizing students in the large variety of levels and making a stressful competition. Now it is an important question when the proposed FDES still evaluating student using infinite-valued logic (fuzzy logic), but the reality is different, and although the FDES system uses infinite-valued logic, it can converge the student's evaluation result to the nearest value of the four values of the current evaluation system (NME, AE, G, and VG) during or at the end of the school year. It means that it uses the infinite-values for considering the real educational abilities of the students but reports the rounded values by four values to avoid stressful competition between them.